\UseRawInputEncoding
\documentclass[journal]{IEEEtran}
\usepackage{cite}

\ifCLASSINFOpdf
  \usepackage[pdftex]{graphicx}
\else
  \usepackage[dvips]{graphicx}
\fi
\usepackage{amsmath}
\ifCLASSOPTIONcompsoc
 \usepackage[caption=false,font=normalsize,labelfont=sf,textfont=sf]{subfig}
\else
 \usepackage[caption=false,font=footnotesize]{subfig}
\fi
\usepackage{url}

\usepackage{xcolor}
\usepackage{booktabs}
\usepackage{algorithm}
\usepackage{algpseudocode}
\usepackage{pifont}
\newcommand{\good}{\ding{51}}      
\newcommand{\excellent}{\ding{51}\ding{51}} 
\newcommand{\bad}{\ding{55}}      

\begin{document}
%
\title{Scaling Fabric-Based Piezoresistive Sensor Arrays for Whole-Body Tactile Sensing}

%
%
%

\author{Curtis C. Johnson,
        Daniel Webb,
        David Hill,
        and Marc D. Killpack
\thanks{The authors are affiliated with the Robotics and Dynamics Laboratory, Department of Mechanical Engineering at Brigham Young University in Provo, Utah, USA}}

%
%

\markboth{Journal of \LaTeX\ Class Files,~Vol.~14, No.~8, August~2015}%
{Shell \MakeLowercase{\textit{et al.}}: Bare Demo of IEEEtran.cls for IEEE Journals}
%



\maketitle

\begin{abstract}
Scaling tactile sensing for robust whole-body manipulation is a significant challenge, often limited by wiring complexity, data throughput, and system reliability. This paper presents a complete architecture designed to overcome these barriers. Our approach pairs open-source, fabric-based sensors with custom readout electronics that reduce signal crosstalk to less than 3.3\% through hardware-based mitigation. Critically, we introduce a novel, daisy-chained SPI bus topology that avoids the practical limitations of common wireless protocols and the prohibitive wiring complexity of USB hub-based systems. This architecture streams synchronized data from over 8,000 taxels across 1 m\textsuperscript{2} of sensing area at update rates exceeding 50 FPS, confirming its suitability for real-time control. We validate the system's efficacy in a whole-body grasping task where, without feedback, the robot's open-loop trajectory results in an uncontrolled application of force that slowly crushes a deformable cardboard box. With real-time tactile feedback, the robot transforms this motion into a gentle, stable grasp, successfully manipulating the object without causing structural damage. This work provides a robust and well-characterized platform to enable future research in advanced whole-body control and physical human-robot interaction.
\end{abstract}


%
\IEEEpeerreviewmaketitle

\section{Introduction}
\label{sec:intro}
\IEEEPARstart{R}{esearch} in robotic manipulation is driven by a desire to enhance the capabilities of robots operating in inherently unstructured environments and manipulating objects of infinite variability. While vision is a powerful modality for robotic manipulation \cite{Kalashnikov_Irpan_Pastor_Ibarz_Herzog_Jang_Quillen_Holly_Kalakrishnan_Vanhoucke_etal._2018}, its utility degrades when objects are occluded or when tasks require more dexterous, force-sensitive interactions. Adding more cameras can mitigate occlusion but does not scale well for complex, open-world scenarios \cite{Cong_Chen_Ma_Liu_Hou_Yang_2023}. In contrast, tactile sensing provides critical information about contact forces, local geometry, textures, and slip that is difficult or impossible to obtain with vision alone, much like haptic feedback improves human manipulation \cite{Bergholz_Ferle_Weber_2023, Dahiya_Metta_Valle_Sandini_2010}.

\begin{figure}[t]
\centering
\includegraphics[width=\columnwidth]{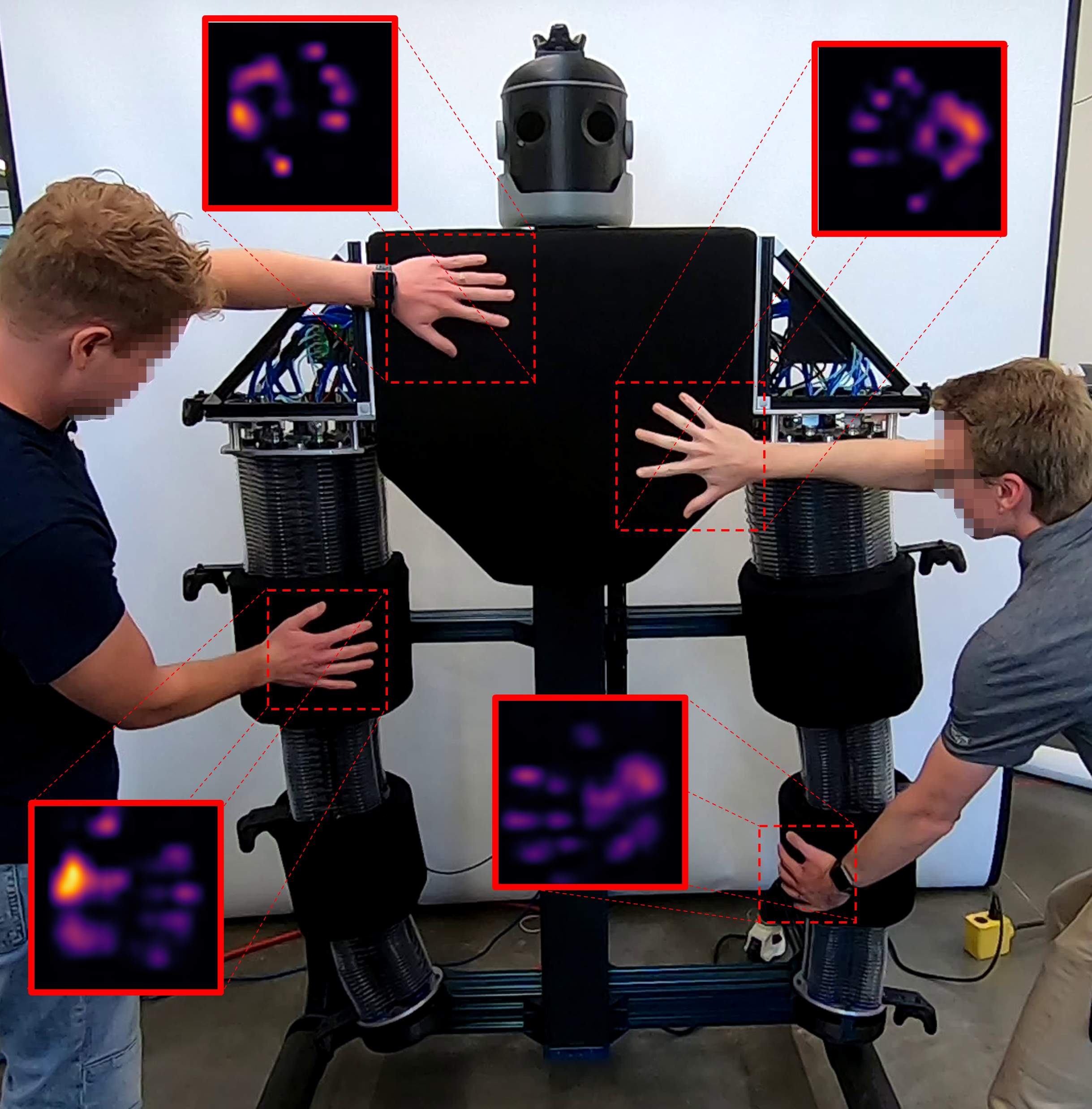}
\caption{Two people pressing on various parts of the robot's sensorized surfaces (in black). This hardware implementation consists of two separate tactile sensing systems with three sensors (3072 taxels), one for each half of the robotic torso (6144 taxels total). All power and communication wiring for the entire system runs down each arm in a single cable channel.}
\label{fig:multiple-hand-touch}
\end{figure}

Historically, robotic tactile sensing has been concentrated at the end-effector, analogous to the human fingertip. This focus has produced a rich variety of transduction methods emphasizing localized, high resolution measurements. Prominent among these are visuo-tactile sensors, such as the GelSight family \cite{Abad_Ranasinghe_2020}, which use cameras to detect deformation in a compliant membrane \cite{Zhang_Chen_Gao_Wan_Shan_Xue_Sun_Yang_Fang_2022, Alspach_Hashimoto_Kuppuswamy_Tedrake_2019, Kuppuswamy_Alspach_Uttamchandani_Creasey_Ikeda_Tedrake_2020, Roberge_Fornes_Roberge_2023}. Other successful approaches include capacitive \cite{Maiolino_Maggiali_Cannata_Metta_Natale_2013, Sarwar_Ishizaki_Morton_Preston_Nguyen_Fan_Dupont_Hogarth_Yoshiike_Qiu_etal._2023, Salim_Lim_2017}, magnetic \cite{Hu-magnetic}, and multi-modal sensors that perceive temperature or slip in addition to pressure \cite{Abad_Reid_Ranasinghe_2022, Mao_Liao_Yuan_Zhu_2024}.

However, humans and animals may often use their entire arm and body to facilitate manipulation of large, heavy, or bulky objects \cite{Dagenais_Hensman_Haechler_Milinkovitch_2021}. Extending sensing from the fingertip to the entire robotic body---a necessity to enable whole-arm or whole-body manipulation---introduces significant scaling challenges that render many end-effector-focuses designs impractical \cite{mandil2023, nilsson2000}. The difficulties are multifaceted and span the physical, electrical, and system levels. Physically, large-scale sensor arrays lead to prohibitive wiring complexity, bulk, and weight from cables, PCBs, and connectors. Electrically, they impose high data throughput requirements, increased power consumption, and greater susceptibility to electromagnetic interference (EMI) over long cables. From a systems engineering perspective, ensuring synchronized, low-latency data acquisition across hundreds or thousands of distributed taxels is critical for real-time control, but difficult to achieve while managing cost and manufacturability \cite{WANG202355}.

\begin{table*}[ht]
\centering
\small
\caption{Qualitative comparison of communication topologies for closed-loop tactile control with three sensors. Ratings reflect practical performance when using three sensors concurrently and are based on empirical measurements and reported behavior in literature.}
\label{tab:combined_topology}
\begin{tabular}{@{}lccccc@{}}
\toprule
\textbf{Topology} & \textbf{Throughput} & \textbf{Low Latency \& Jitter} & \textbf{Scalability} & \textbf{Minimal Wiring} & \textbf{EMI Robustness} \\
\midrule
I\textsuperscript{2}C Bus \cite{Fiedler_Ruppel_Jonetzko_Hendrich_Zhang_2021, Schmitz_Maiolino_Maggiali_Natale_Cannata_Metta_2011}        & \bad        & \bad        & \bad        & \good       & \good \\
Peer-to-Peer \cite{Mittendorfer_Yoshida_Cheng_2015, Shen_Armleder_Cheng_2025}                      & \bad        & \bad        & \bad        & \bad        & \good \\
Star (Bluetooth) \cite{Fiedler_Ruppel_Jonetzko_Hendrich_Zhang_2021, Murphy_Zhu_Liang_Matusik_Luo_2024}                 & \bad        & \bad        & \excellent  & \excellent  & \bad  \\
Star (Wi-Fi) \cite{Murphy_Zhu_Liang_Matusik_Luo_2024}                      & \good       & \bad        & \excellent  & \excellent  & \bad  \\
Star (USB 2.0 Hub) \cite{Luo_Li_Sharma_Shou_Wu_Foshey_Li_Palacios_Torralba_Matusik_2021}               & \excellent  & \good       & \good       & \bad        & \good \\
\textbf{SPI + USB 2.0 (Ours)}     & \good  & \good       & \excellent  & \good       & \good \\
\bottomrule
\end{tabular}
\end{table*}

Fabric based piezoresistive sensors offer a compelling path forwards for large-area sensing due to their inherent low cost, durability, ease of manufacturing, light weight, and good sensitivity \cite{Saxena_Patra_2024, Luo_Li_Sharma_Shou_Wu_Foshey_Li_Palacios_Torralba_Matusik_2021}. These sensors have been demonstrated in arrays with hundreds \cite{Luo_Li_Sharma_Shou_Wu_Foshey_Li_Palacios_Torralba_Matusik_2021} to thousands of taxels \cite{Warnakulasuriya_Dinushka_Dias_Ariyarathna_Ramraj_Jayasinghe_DeSilva_2021}. Yet they are not without their own challenges, most notably electrical crosstalk between sensing elements in an array \cite{dominguez2023, Day_Penaloza_Santos_Killpack_2018} and material property variations inherent to fabrics and polymers \cite{ma15155185}. To address these, our readout electronics, presented on a compact, credit-card-sized PCB, implement the zero potential grounding method to substantially reduce crosstalk \cite{Warnakulasuriya2021} and incorporate an adjustable sensitivity to accommodate material property inconsistencies \cite{Murphy_Zhu_Liang_Matusik_Luo_2024}.

Crucially, a scalable sensor technology must be paired with a scalable system architecture. Whole-body systems must provide reliable, synchronized measurements to support closed-loop control during unpredictable, distributed contact events \cite{Mittendorfer_Yoshida_Cheng_2015, murooka2024, jiang2024hierarchical}. We target a minimum frame rate of 50 FPS to ensure sufficient temporal resolution for closed-loop tactile control. Achieving this frame rate requires a data throughput of at least 1 Mbps per sensor. This target is informed by neurophysiological findings that human perception of tactile frequency saturates near 50 Hz, suggesting this is a sufficient rate for conveying rich temporal feedback for manipulation tasks \cite{Graczyk_Christie_He_Tyler_Bensmaia_2022}.

A critical challenge in scaling up tactile systems is the choice of communication topology, which must balance bandwidth and latency with the practical constraints of physical integration. As summarized in Table~\ref{tab:combined_topology}, many common architectures are ill-suited for large-scale robotic applications. Peer-to-peer topologies, for instance, become prohibitive in terms of cost and weight; a system with several thousand taxel PCBs each weighing 3 grams could add several kilograms to a robot \cite{Mittendorfer_Yoshida_Cheng_2015}, severely limiting its manipulation capabilities (e.g., the commercially-available Franka Emika Panda robot has a maximum payload of 3 kg). Star topologies also present significant issues: wireless systems like Wi-Fi or Bluetooth are susceptible to interference and lack deterministic performance without specialized, non-standard hardware \cite{realtime-wifi}. Wired USB hubs provide excellent performance but become unwieldy with extensive cabling that creates multiple points of failure. While a shared hardware bus offers a more robust foundation, the commonly used I\textsuperscript{2}C protocol is severely constrained by a low data rate (up to 400 kbps) and susceptibility to EMI over long cable runs. In contrast, our shared SPI bus architecture provides a superior solution, offering a data rate over 100 times faster (42 Mbps) and with greater noise immunity, enabling a robust and scalable communication backbone for real-time control.

This paper presents a complete, scalable solution for whole-body tactile sensing that addresses challenges at both the sensor and system level. Our main contributions are:

\begin{itemize}
\item An open-source, fabric-based sensor design and drive electronics\footnote{\url{https://github.com/byu-rad-lab/tactile-sensor}\label{repo}} that integrate best-practice crosstalk reduction circuitry and tunable sensitivity for robust performance.
\item A novel, high-throughput shared bus architecture using SPI that ensures deterministic timing and system reliability for distributed sensor arrays.
\item A full-system demonstration on a robotic torso with soft, air-driven continuum joints, showcasing the system's efficacy for closed-loop, whole-body manipulation tasks that are difficult to achieve with other sensing architectures.
\end{itemize}

The remainder of this paper is organized as follows. Section \ref{sec:sensor-design} and \ref{sec:pcb-design} discuss the design of the sensor and drive electronics. Section \ref{sec:system-architecture} details the system architecture. Section \ref{sec:experiments} and \ref{sec:results-discussion} present our experimental methods and results. Finally, Section \ref{sec:conclusion} provides a summary and directions for future work.

\section{Methods}
\subsection{Sensor Design}
\label{sec:sensor-design}
The tactile sensors used in this work are based on the fabric piezoresistive array design first introduced in \cite{bhattacharjee2013tactile, Day_Penaloza_Santos_Killpack_2018}. While the literature contains many variations, from knitted textiles \cite{Luo_Li_Sharma_Shou_Wu_Foshey_Li_Palacios_Torralba_Matusik_2021} to arrays using conductive foils with piezoresistive polymers like Velostat \cite{Fiedler_Ruppel_Jonetzko_Hendrich_Zhang_2021}, we selected a fabric-based approach for its superior flexibility and durability under large, unpredictable loads. Foil-based sensors, in contrast, can be prone to cracking under the repeated bending inherent to whole-body robotic applications. As the sensor fabrication is not a novel contribution of this paper, this section provides a brief overview of the design; we refer readers to \cite{li2020recent} and \cite{chen2020progress} for a broader survey of sensor designs for various applications.

\begin{figure}
\includegraphics[width=0.9\columnwidth]{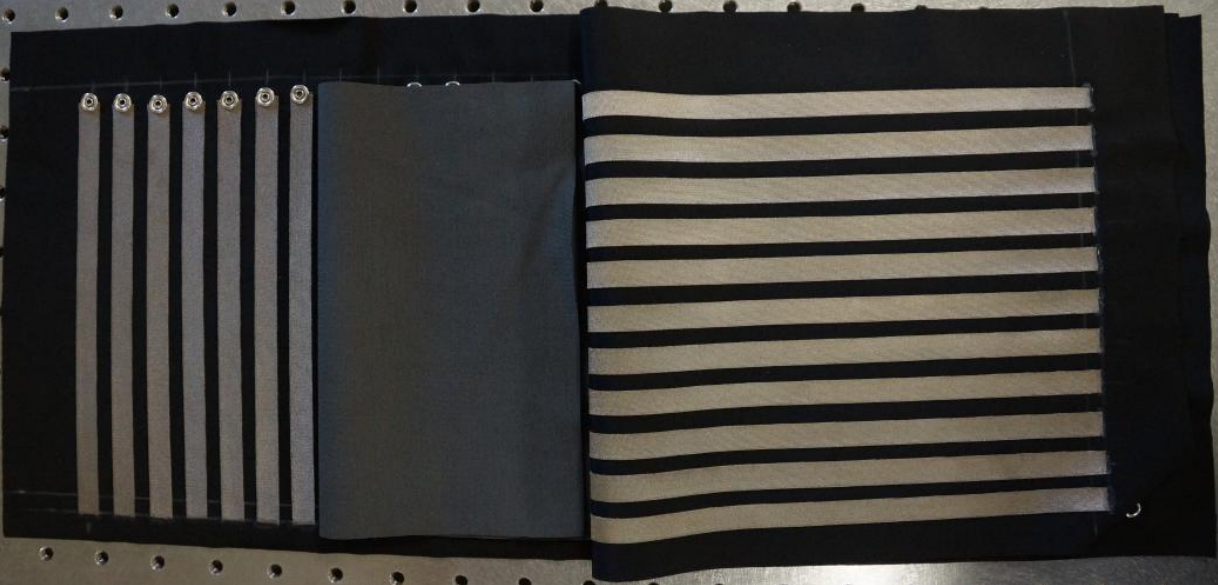}
\centering
\caption{The layers of the fabric tactile sensor, shown unfolded. From outside in: non-conductive spandex substrate (black), parallel conductive fabric strips (light gray), and a central piezoresistive fabric layer (dark gray).}
\label{fig:fabric_sensor}
\end{figure}

As illustrated in Figure \ref{fig:fabric_sensor}, the sensor consists of a piezoresistive fabric layer sandwiched between two orthogonal sets of parallel conductive strips. These strips are adhered to a flexible, non-conductive spandex substrate. This construction forms a grid where each intersection acts as a tactile pixel, or `taxel.' The conductive strips are connected to a ribbon cable via metallic snaps, which interfaces with the drive electronics for data acquisition.
During operation, the volumetric resistance of the piezoresistive material at each taxel, denoted as $R_{i,j}$ in Figure \ref{fig:peripheral-block-diagram}, decreases as a function of applied pressure. The drive electronics scan the array by applying a known voltage to each row sequentially and measuring the resulting current from each column, yielding a voltage signal proportional to the pressure at each taxel location.

\subsection{Peripheral PCB Design}
\label{sec:pcb-design}

The sensor readout electronics are implemented on a compact, credit-card-sized peripheral PCB, shown in Figure~\ref{fig:pcbs}. To achieve the deterministic timing required for closed-loop control, our design intentionally avoids a microcontroller-based architecture, which can introduce timing variability from software loops, interrupt handling, and peripheral latency. While an FPGA \cite{oballefpga} could offer similar determinism, our approach using discrete digital logic and a dedicated SPI-enabled Analog-to-Digital (ADC) provides comparable timing guarantees at a significantly lower cost and design complexity. Furthermore, this hardware-based approach eliminates the need for any peripheral-specific firmware which simplifies system deployment. 

We organize the readout electronics as shown in Figure~\ref{fig:peripheral-block-diagram}, with an input module responsible for applying voltages to a specific sensor input and an output module responsible for reading voltages from the sensor. Each module has specific crosstalk mitigation circuitry that will be discussed in its own section. 

Generally, each peripheral board is controlled by a central controller via an 8-wire bus carrying signals for Chip Select (CS), SPI communication (DATA, CLK), counter control (CNT, CLR), and power (3.3V, GND). The shared signals are routed over standard RJ45 connectors and ethernet cables, allowing multiple peripherals to be daisy chained. The CS signal is sent on a dedicated twisted pair to allow unique addressing of each board.

\subsubsection{Input Module}

The input module selectively applies a known voltage to one of the 16 sensor inputs.  It uses a ring counter, comprised of a D-type flip-flop (Texas Instruments SN74LVC1G74DCUR) and two 8-bit shift registers (Texas Instruments CD74AC164M96), to circulate a single digital high signal. This signal activates a low on-resistance SPDT switch (Analog Devices ADG839YKSZ-REEL7) corresponding to the selected input, connecting it to the 3.3V power rail. All other SPDT switches remain in their normally closed state, connecting the inactive inputs to ground. This design choice is critical for crosstalk mitigation, as detailed in Section \ref{sec:pcb-design}. Each pulse of the CNT signal advances the active input channel forward by one and the CLR signal resets the counter. The ultra-low 0.35 $\Omega$ on-resistance of the switches ensures a negligible voltage drop on the selected input channel.

\subsubsection{Output Module}

The output module sequentially connects each of the 64 sensor output channels to the data acquisition (DAQ) circuitry. This is achieved using a 64:1 multiplexer stage, built from four 16-channel multiplexers (Analog Devices ADG706BRUZ-REEL7) whose outputs are fed into a 4-channel multiplexer (Analog Devices ADG804YRMZ-REEL). A 12-bit binary counter (Texas Instruments SN74LV4040APWR), synchronized with the input module via the CNT and CLR signals, coordinates the output channel selection. This ensures that all multiplexer channels are switched together to deliver one sensor output to the ADC at a time.

We selected these multiplexers for their low on-resistance (0.5 $\Omega$ and 2.5 $\Omega$) and excellent on-resistance flatness (0.1 $\Omega$ and 0.5 $\Omega$). While the on-resistance introduces a small parasitic resistance which affects the signal we are trying to measure, its flatness ensures this effect is uniform across channels and can be compensated during calibration. 

The DAQ circuitry implements the zero-potential (or virtual ground) method. An operational amplifier (Analog Devices LT1807CS8\#PBF) in a transimpedance configuration holds the selected column at a virtual ground, converting the incoming current from the taxel into a voltage. A second amplifier inverts this signal to match the input range of the ADC.  The resulting function relating input to output voltages is

\begin{equation}
    V_{\text{out}} = \frac{R_f}{R_{i,j}} V_{\text{in}}
\end{equation}

\noindent where $V_{\text{in}} = 3.3V$ $R_{i,j}$ is the resistance of the active taxel, and $R_f$ is an adjustable feedback resistor that sets the system sensitivity. The rail-to-rail op-amp ensures the virtual ground is held close to zero potential because of its high open-loop gain (120 dB) and low input offset voltage ($<$ 50 $\mu$V).

Data acquisition is handled by a 12-bit SPI-enabled ADC (Maxim Integrated MAX11103AUB+). When the controller pulls the peripheral board's CS line low, the ADC performs a conversion and transmits the data over the shared SPI bus. The acquisition speed is determined by the SPI clock, which was set to 14 Mhz for all experiments reported in this paper. The choice of this speed is affected by how far the signal must travel, sources of signal interference, and desired throughput. For our work, 14 Mhz provided a good balance between throughput and signal integrity over $\approx$1.3 meters of signal travel.

\begin{figure}[t]
\centering
\includegraphics[width=\columnwidth]{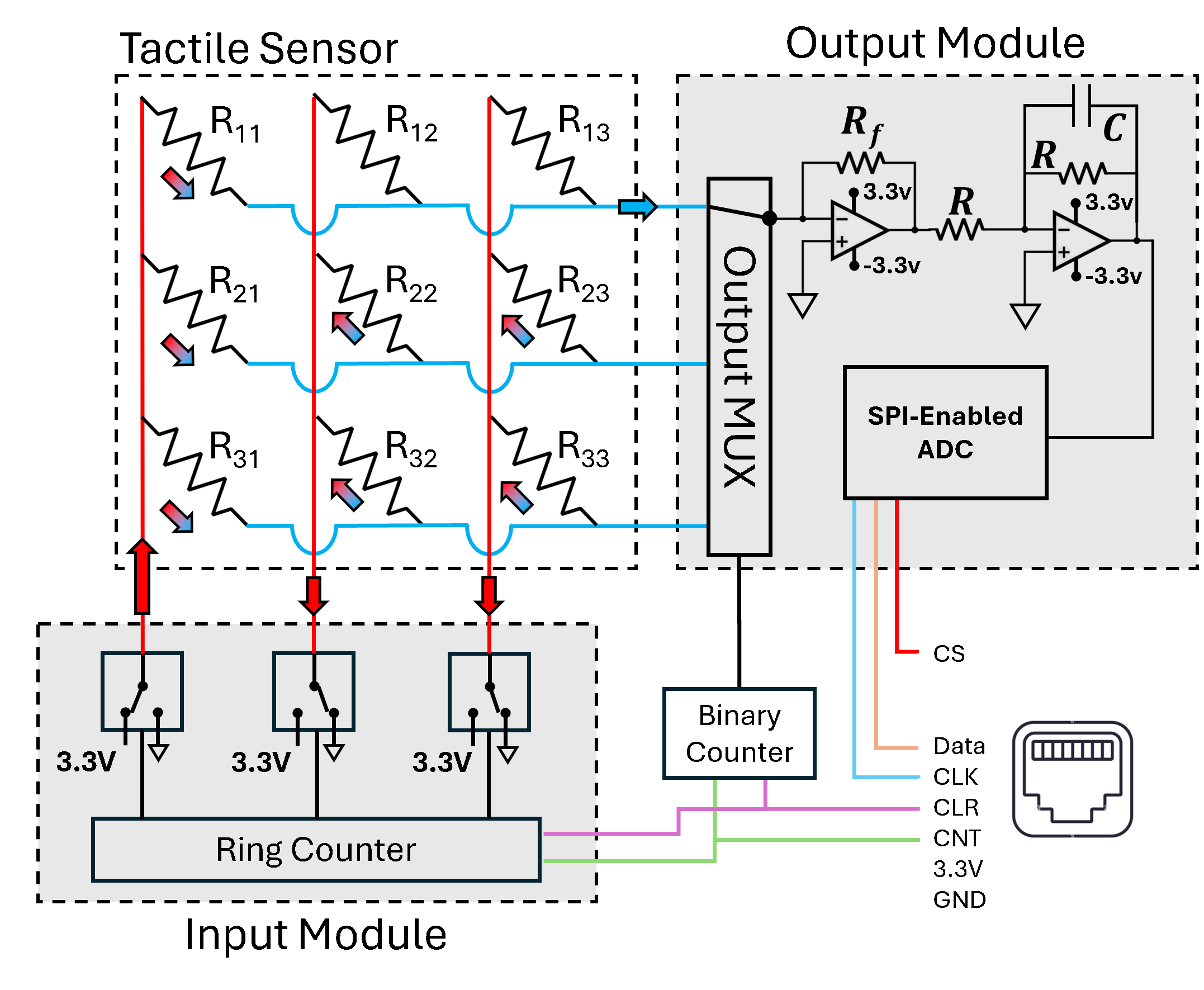}
\caption{Block diagram of the sensor readout peripheral, configured to measure taxel $R_{1,1}$. The input module activates row 1, while the 64:1 output multiplexer stage selects column 1 for measurement by the DAQ circuitry. Red indicates input channels and blue indicates output channels.}
\label{fig:peripheral-block-diagram}
\end{figure}

\subsubsection{Crosstalk Mitigation}

Crosstalk is minimized through a two-fold approach at both the input and output stages of the circuit.

The primary mechanism is the grounding of all non-active sensor rows by the input module's SPDT switches. This technique shunts parasitic current paths to ground before they can interfere with the measurement on the active column. Crucially, this is implemented with dedicated analog switches rather than microcontroller GPIO pins. The transistors in GPIOs often fail to provide a true rail-to-rail output, resulting in load-dependent and non-zero ground potentials on the unselected input channels. Our use of low on-resistance analog switches ensures a stable, near-zero potential on all inactive rows.

At the output stage, the transimpedance amplifier used to implement the virtual ground method holds the selected output channel very close to zero potential. This ensures that all of the current flowing through the selected taxel flows into the output module. 

The combination of these two methods effectively mitigates the main parasitic current paths. To illustrate, consider the measurement of the voltage drop across $R_{1,1}$ as shown in Figure~\ref{fig:peripheral-block-diagram}. Input 1 is driven high, while all other input rows are actively grounded. Current flows through Input 1 to the sensor. This current is divided among $R_{1,1}$, $R_{2,1}$ and $R_{3,1}$ and flows down to the output channels. Current is free to `back flow' from Outputs 2 and 3 to Inputs 2 and 3, as indicated by the reversed arrows. Importantly, all of the back flow current is shunted back to ground through the unselected inputs. Only very small parasitic currents flow through $R_{1,2}$ or $R_{1,3}$ because both sides of the resistors are kept very close to zero potential: one side by the input SPDT switches and the other side by the virtual ground method.

The efficacy of this mitigation strategy is fundamentally limited by the non-zero on-resistance of the analog switches and multiplexers. These small resistances mean the ground potentials are not perfectly zero, leading to minor, load-dependent voltage drops. As the total current through the system increases (e.g., when many taxels are pressed simultaneously), these voltage drops can become more pronounced, allowing for small amounts of residual crosstalk. However, for most manipulation tasks, this effect is negligible. We perform an experiment quantifying this effect in Section ~\ref{sec:sensor-validation}.

\subsection{System Architecture}
\label{sec:system-architecture}

The system architecture is a core contribution of this work and is designed to enable robust, scalable, and synronized whole-body tactile sensing. This section justifies our selection of the shared SPI bus topology over alternative methods and then presents the details of the hardware implementation, communication protocols, and signal integrity measures.

\subsubsection{Communication Topology}
As established in Section~\ref{sec:intro}, high-throughput topologies are required for real-time tactile control. Wi-Fi, USB Hub, and SPI (see Table~\ref{tab:combined_topology}) can all theoretically meet our bandwidth requirements but present different tradeoffs for practical deployment on a physical robot: 

\begin{itemize}
    \item \textbf{Wi-Fi}: The primary appeal of a wireless topology is the complete elimination of physical cables. This is a significant advantage for robots with high degrees of freedom or when sensors must be separated by large distances. Despite its potential, wireless systems face significant hurdles for reliable, real-time control. Novel methods to address Wi-Fi's inherent latency and jitter remain largely conceptual and are unsupported by any commercially-available hardware \cite{realtime-wifi}. Furthermore, the high-EMI environment of a robot with multiple actuators makes wireless communication prone to interference and dropouts. Given these reliability concerns and the lack of suitable hardware, a direct experimental comparison was not feasible, rendering this approach unsuitable for our application.
    \item \textbf{USB Hub}: A star topology using USB hubs is a strong candidate due to its high bandwidth, low latency, and inherent parallelism. Because each device can communicate with the host independently, adding more sensors does not affect the frame rate until the hub bandwidth is saturated. In terms of throughput, a USB hub is highly scalable. However, this performance comes at the cost of physical scalability. As shown in Table~\ref{tab:topology_comparison_scaling}, each sensor requires a separate long cable run to a central hub, which increases wiring complexity, weight, cost, power draw, risk of noise-inducing ground loops, and the number of potential mechanical points of failure. For our robotic platform, the physical volume required to route several USB cables was simply unavailable, making this topology infeasible for our application.
    \item \textbf{SPI}: Our daisy-chained SPI bus provides the best balance of performance and practical deployability for our application. While both SPI and USB can meet our performance targets, they scale differently. In contrast to the parallel nature of USB, our SPI bus operates serially, and thus the frame rate decreases as more sensors are added (Table~\ref{tab:topology_comparison_scaling}). Despite this limitation, our design was chosen because it easily meets our 50 FPS requirement at the intended scale while offering significantly lower wiring complexity, reduced system weight and power draw, and better noise immunity in our high-EMI environment. 
\end{itemize}

\begin{table}
\caption{System-level scaling comparison of communication topologies as the number of sensor nodes ($N$) increases.}
\label{tab:topology_comparison_scaling}
\centering
\begin{tabular}{@{}lcc@{}}
\toprule
\textbf{System-Level Metric} & \textbf{Shared SPI Bus (Ours)} & \textbf{USB Hub} \\ 
\midrule
Wiring Complexity & Low Increase (\(\uparrow\)) & High Increase (\(\uparrow \uparrow\)) \\
System Weight     & Low Increase (\(\uparrow\)) & High Increase (\(\uparrow \uparrow\)) \\
System Power Draw & Low Increase (\(\uparrow\)) & High Increase (\(\uparrow \uparrow\)) \\
System Cost       & Low Increase (\(\uparrow\)) & High Increase (\(\uparrow \uparrow\)) \\
Risk of EMI / Ground Loops & Constant ($=$) & High Increase (\(\uparrow \uparrow\)) \\
Frame Rate (per sensor) & Decreases (\(\downarrow\)) & Constant ($=$) \\
\bottomrule
\end{tabular}
\end{table}

\subsubsection{Hardware Implementation}
\begin{figure}
    \includegraphics[width=.7\columnwidth]{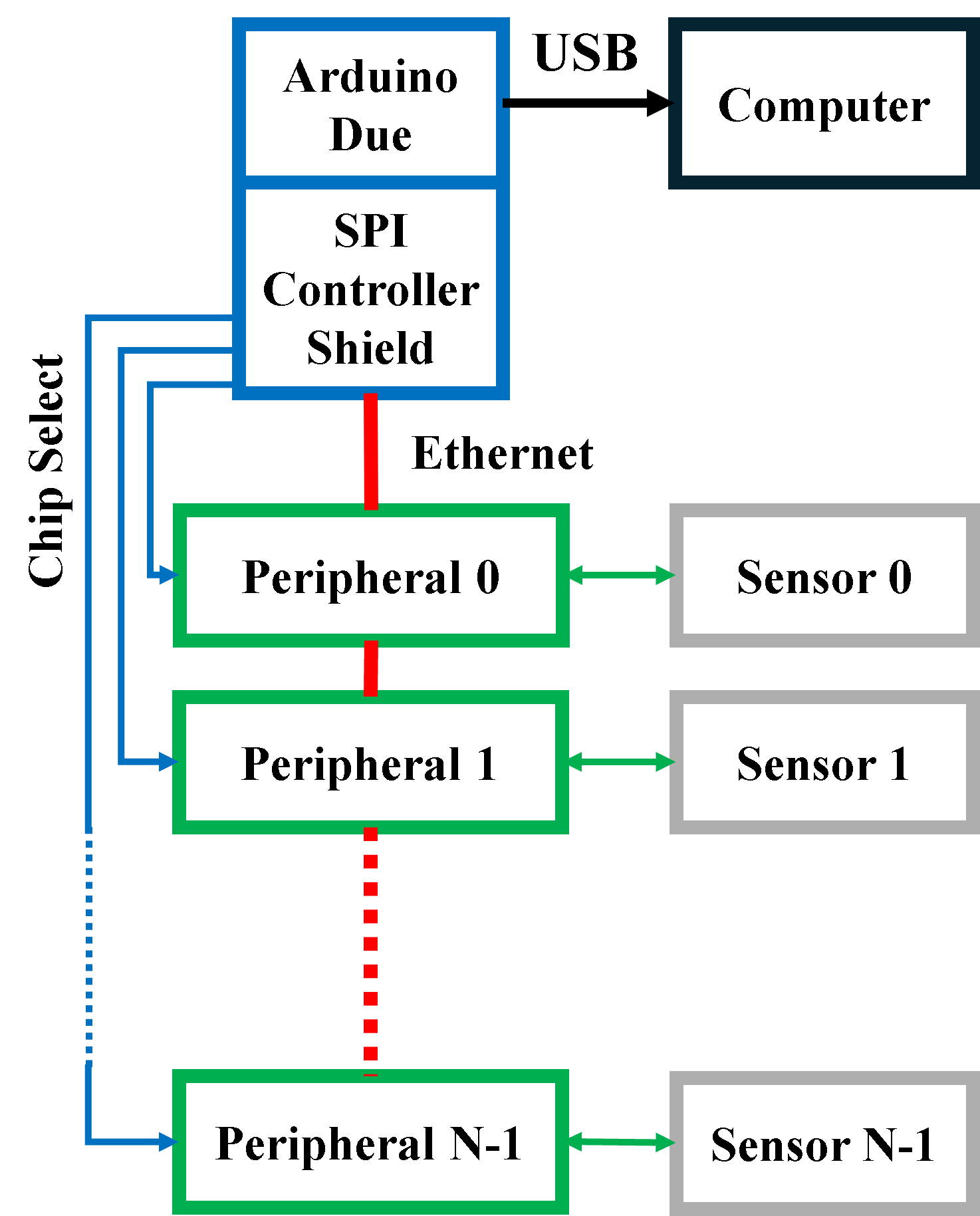}
    \centering
    \caption{Tactile sensor PCB block diagram.}
    \label{fig:system-block-diagram}
\end{figure}

The system hardware is comprised of a central controller that orchestrates communication with a series of daisy-chained peripheral boards, as shown in Figure~\ref{fig:system-block-diagram}. 

\begin{figure}
    \centering
    \includegraphics[width=0.8\columnwidth]{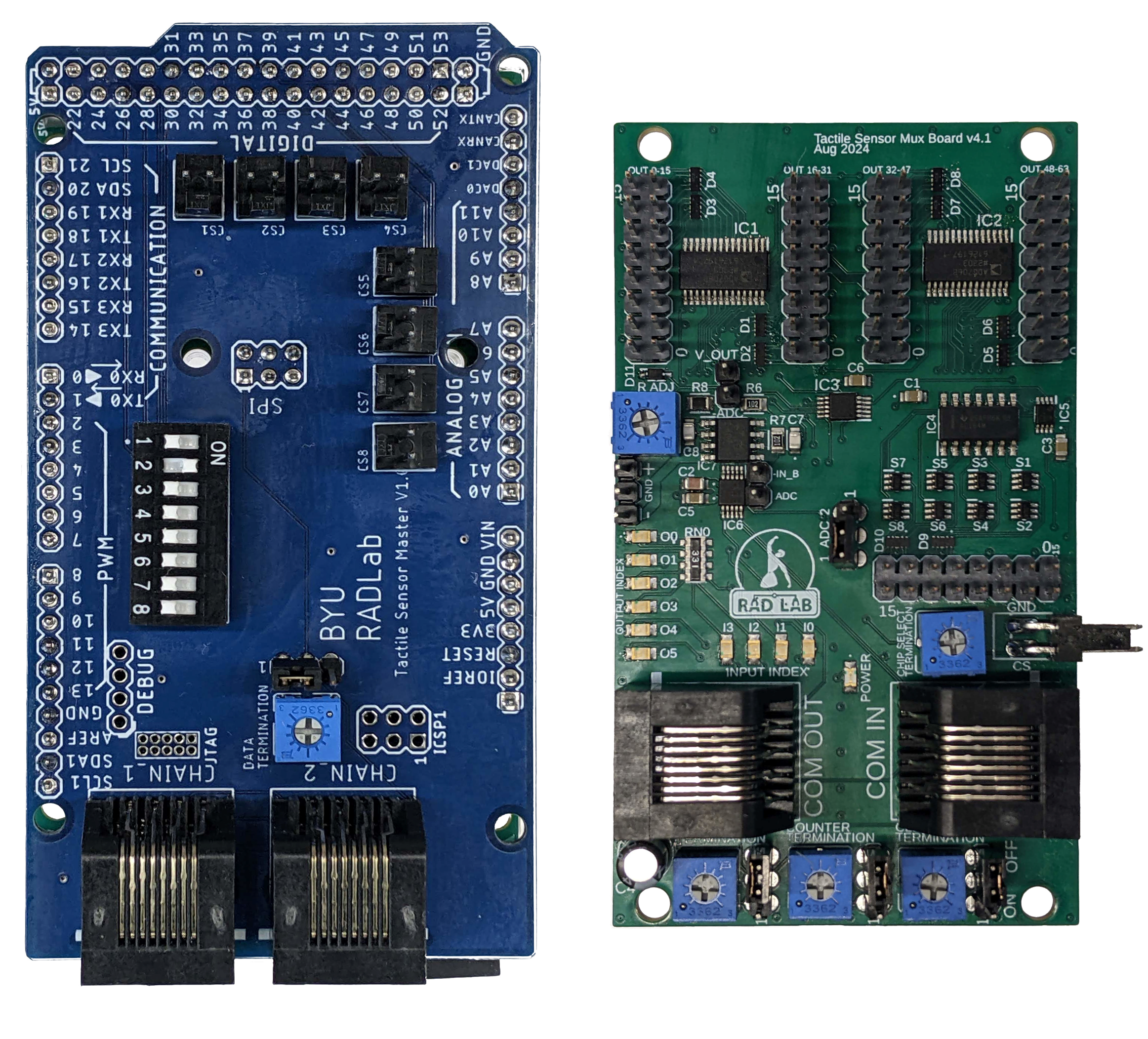}
    \caption{SPI Controller Shield (Left) and sensor readout peripheral pcb (right).}
    \label{fig:pcbs}
\end{figure}

The controller is an Arduino Due with a custom designed shield (Figure~\ref{fig:pcbs}, left). We selected the Due specifically for its native USB port, which provides high-speed serial communication (up to 480 Mbps theoretical; about 70 Mbps observed in practice). This is essential for transmitted large aggregated data packets to the host computer. The custom shield serves as the physical interface to the sensor chain. It includes a dip switch to allow the user to specify the number of peripherals on the bus to dynamically preallocate the correct buffer size and optimize efficiency. 

The peripheral daisy chain consists of one or more sensor readout PCBs. Each peripheral board (Figure~\ref{fig:pcbs}, right) connects to a single tactile sensor of any size up to a maximum of 16x64 taxels and implements the electronics discussed in Section~\ref{sec:pcb-design}. The controller shield supports up to 8 peripherals, enabling a single controller to manage a system of up to 8,192 unique taxels. For systems requiring even higher frame rates, multiple independent controllers can be used in parallel.

We prioritized signal integrity and system protection to ensure robust operation in a real-world robotic environment. To manage the signal reflections and noise that can occur when transmitting high-frequency digital signals over long cable runs (up to 2m), we implemented a specific termination strategy. We leverage the 120$\Omega$ characteristic impedance of the Cat5 cables and include potentiometers on each peripheral board that function as adjustable termination resistors. For the shared bus signals (SPI CLK, CNT), we adjust the termination resistor on the last board in the chain to match the 120$\Omega$ impedance, which absorbs signal energy and prevents reflections. For dedicated point-to-point signals like the CS line, we use the termination resistor on each individual peripheral. As a complementary measure, we added small-value capacitors to the data and clock lines to slightly reduce the signal slew rate, which further suppresses high-frequency ringing. This careful termination strategy ensures clean signals throughout the entire chain. Furthermore, we equipped all external-facing ports on the shield and peripherals with ESD protection diodes to guard against electrostatic discharge which occurs frequently when interacting with various materials and objects during whole-body manipulation.

\subsubsection{Data Acquisition}

The process for acquiring data from all the sensors is managed by the controller which transmits the data to the host computer via high-speed USB.

The algorithm running on the SPI controller is summarized in Algorithm~\ref{alg:readout}. The process begins with a pulse on the CLR line, which resets all the counters on the bus to the first taxel (input 0, output 0). The controller then iterates through every taxel on the sensor. For each taxel coordinate, the controller selects a peripheral ($N$) in the daisy chain by pulling its CS line low, initiating an ADC conversion and data transfer. After every peripheral has been queried for the current taxel coordinate, the controller pulses the CNT line to advance all the peripherals to the next taxel in unison. A brief 1 $\mu$s delay is included after each CNT pulse to allow the analog signals to settle before the next measurement. This delay time is dictated by the rise time of the low pass filter $R$ and $C$ in the output module (Figure~\ref{fig:peripheral-block-diagram}).

\begin{algorithm}
\caption{Tactile Sensor Readout over SPI Bus \label{alg:readout}}
\begin{algorithmic}[1]
\State Pulse \texttt{CLR} pin to reset taxel counter
\For{$i = 0$ to $N_{\text{out}} - 1$} \Comment{For each output line}
    \For{$j = 0$ to $N_{\text{in}} - 1$} \Comment{For each input line}
        \For{$k = 0$ to $N - 1$} \Comment{For each sensor}
            \State Set \texttt{CS}[$k$] LOW \Comment{Start transmission}
            \State 16-bit read from ADC 
            \State Set \texttt{CS}[$k$] HIGH \Comment{End transmission}
        \EndFor
        \State Pulse \texttt{CNT} pin \Comment{Advance to next taxel}
        \State Wait $1\,\mu$s \Comment{Settling time for analog filter}
    \EndFor
\EndFor
\State Format all data into serial message \Comment{see Table~\ref{tab:message-format}}
\State Transmit formatted message to host via USB
\end{algorithmic}
\end{algorithm}

The total time required to complete one full scan of the entire system is deterministic and described by

\begin{equation}
t_{\text{total}} = N_{\text{out}} \times N_{\text{in}} \times \bigg[ N \times \left(t_{\text{SPI}} + t_{\text{proc}}\right) + t_{\text{delay}} \bigg]
\label{eq:runtime}
\end{equation}

In our system, $t_{\text{SPI}}$ is the time for a 16-bit transfer over the 14 Mhz SPI bus ($\approx 1.14 \mu$s), $t_{\text{proc}}$ is the controller's processing overhead per peripheral ($\approx 1.08 \mu$s) and $t_{\text{delay}}$ is the fixed settling time ($1 \mu$s). This equation clearly shows that the total scan time, and thus the frame rate, scales linearly with the number of peripherals on the bus $N$. Consequently, the system frame rate (1/$t_{\text{total}}$ has a non-linear decay as more sensors are added to the chain. 

On the host computer, we used the PySerial python package to read the data from the Arduino Due and then publish the data via ROS. The data for the entire system arrives in a packet formatted like Table~\ref{tab:message-format}. Before publishing, we filtered the data using a Hampel filter on a time-series history of each taxel. The Hampel filter is a robust outlier detection algorithm that specializes in removing spurious noise spikes which can arise from EMI, without distorting the rest of the signal. For a rolling window of recent measurements for a given taxel, it calculates the median and median absolute deviation (MAD). IF the newest data point is more than a set number of standard deviations (typically 2-3) from the median, it is replaced by the median value. To avoid computational bottlenecks during filtering, we used the Numba python library to just-in-time (JIT) compile this function.

\begin{table}
\caption{USB Serial Message Format}
\label{tab:message-format}
\centering
\begin{tabular}{@{}lll@{}}
\toprule
\textbf{Field}           & \textbf{Size (bytes)} & \textbf{Description}                                                        \\ \midrule
Start Byte               & 2                    & 0xFFFF                   \\
Number of Sensors        & 2                    & Number of active sensors $N$                        \\
Sensor IDs               & $N$             & List of active sensor IDs                             \\
Sensor Data              & $(1024 \times 2) \times N $             & ADC values              \\
\bottomrule
\end{tabular}
\end{table}

 \subsection{Experimental Methods}
\label{sec:experiments}

This section describes the various experimental methods used to 1) quantify the performance of our proposed system, 2) validate the crosstalk mitigation and adjustable sensitivity designs of the readout electronics, and 3) demonstrate the importance of whole-body tactile feedback with an ablation study grasping a large box. 

\subsubsection{System Performance}
We conducted a series of experiments to quantify the key performance metrics of our system architecture. Specifically, we characterized its data throughput and frame rate, analyzed the computational load on the host computer, and measured its end-to-end latency and jitter to validate its suitability for real-time control.

\paragraph{Frame Rate vs Number of Sensors}
This experiment quantified the relationship between the system's frame rate and the number of connected sensors ($N$) and the SPI bus speed. We varied $N$ from 1 to 8 sensors and the SPI clock from 7 Mhz to 42 Mhz and measured the average wall-clock time required for a full acquisition cycle (Algorithm~\ref{alg:readout}). This experiment empirically validated the scaling behavior predicted by Equation~\ref{eq:runtime}.

\paragraph{CPU Load vs Number of Sensors}
To investigate the computational load on the host system and to identify the primary performance bottleneck, we measured the CPU utilization of individual threads within the main ROS process. Using the psutil library, we distinguished between the main application thread (identified by its thread ID matching the process ID) and the busiest worker thread, which corresponds to the pyserial I/O handling thread. For each sensor configuration from $N=1$ to $N=8$, we monitored the cumulative CPU time of each thread over a 60-second period to calculate the average CPU utilization percentage for both the main and the reader threads. This experiment allowed us to precisely quantify how the computational burden shifts from I/O handling to data processing as the number of sensors increased. This relationship is particularly important for deployment on embedded computers with potentially limited CPU processing power. 

\paragraph{End-to-End Latency and Jitter}
A low and predictable end-to-end latency is critical for effective closed-loop control. We designed this experiment to quantity this metric for our system during a dynamic task that mimics real-world contact detection scenario. We conducted this experiment with three sensors integrated into the robot arm, with the SPI clock running at 14 Mhz. 

To generate a repeatable impact event, we commanded the robot arm to swing until its integrated tactile sensor made contact with a stationary reference Force-Torque (FT) sensor. Upon detection of a force spike by the tactile system, a stop command was immediately issued to the robot's controller.

For latency analysis, we used the ROS message header timestamp, synchronized to the host machine using Chrony, as a common time base. The computers are all hard wired together, so the average clock offsets was on the order of $\approx$ 200 ns. Due to the different sampling rates of the sensors (300 Hz for the FT sensor, 120 Hz for the tactile sensor), we identified corresponding events by finding the timestamps of peak signal magnitudes during the impact. The difference in timestamps between the peak of the FT sensor and that of the tactile sensor is the latency. This peak comparison method has the advantage of being invariant to differences in signal magnitudes between the two sensors. 

We repeated this process 50 times to gather statistics on the latency measurements. For additional context to understand the jitter measurements, the primary source of measurement uncertainty is due to timestamp quantization, which is approximately $\pm$ 4.2 ms for the tactile sensor and $\pm$ 1.7 ms for the FT sensor for a combined uncertainty of $\pm$ 4.5 ms (using root sum of squares). This experiment validates the systems responsiveness during closed-loop tactile control.  

\subsubsection{Sensor Validation}
\label{sec:sensor-validation}
We conducted two experiments to validate the performance and key features of the individual sensor and peripheral board design.

\paragraph{Crosstalk Measurement}
This experiment quantifies the efficacy of our hardware-based crosstalk mitigation techniques discussed in Section~\ref{sec:pcb-design}. We designed this test to measure the sensor's susceptibility to `ghosting' or `phantom force' artifacts, which can occur when multiple taxels are activated simultaneously, causing parasitic current paths to form. To create a worst case scenario, we forced three separate taxels to saturate by pressing on them until the corresponding voltage output stopped responding (i.e., railed high). While the taxels are all pressed, we recorded the voltage output of a fourth untouched taxel affected by the parasitic current paths. The crosstalk is the voltage reading on the `ghost' taxel, expressed as a percentage of the full scale saturation voltage. 

\paragraph{Adjustable Gain Demo}
To demonstrate the practical utility of the adjustable gain feature for tuning the sensor's sensitivity, we conducted a simple validation using a single static weight placed on a sensor. While continuously recording the sensor output, we manually adjusted the on-board potentiometer through its full range and record the sensor's output over time. This feature can help compensate for variations in fabric materials or optimizing the sensor's dynamic range for a specific task.

\subsubsection{Whole-Arm Tactile Sensing Ablation Study}
Using a predefined pressure trajectory for the pneumatic continuum arms shown in Fig.~\ref{fig:multiple-hand-touch}, we grasped a large box. Our goal for this experiments was to 1) demonstrate full-body tactile sensing system and 2) assess whether the sensor can be used to prevent object damage during a grasp that would otherwise result in crushing.

The robot executes a predefined grasping trajectory through a set of commanded joint pressures. First we followed the trajectory open-loop (i.e. without tactile feedback), observing the resulting deformation of the boxes. Then we follow the same trajectory, but modulate it according to tactile feedback, and observe the differences. Because torques are caused by antagonistic pressure differentials in the joints, we modulated the trajectory by subtracting from the pressure that causes bending in the grasp direction ($p^+$) from the joint preceding the tactile sensor (e.g., the first joint in response to sensor R1 -- see Figure~\ref{fig:grasp-comparison}). This results in a lower pressure differential and less bending in that direction. The general form of this operation is 

\begin{equation}
    p^+_{\text{cmd}} = p^+_{\text{nominal}} - k_p \sum V_{\text{sensor}}
\end{equation}

\noindent where $\sum V_{\text{sensor}}$ is the sum of the ADC values over the whole sensor and $k_p$ is a proportional gain. Though active during the experiments, we did not use modulate pressures based on the activations of the chest sensors. 

We evaluated the condition of the object after each grasp qualitatively and record the outcomes of open-loop and closed-loop conditions.

\section{Results and Discussion}
\label{sec:results-discussion}

\subsection{System Performance}
\begin{figure}
    \centering
    \includegraphics[width=0.9\columnwidth]{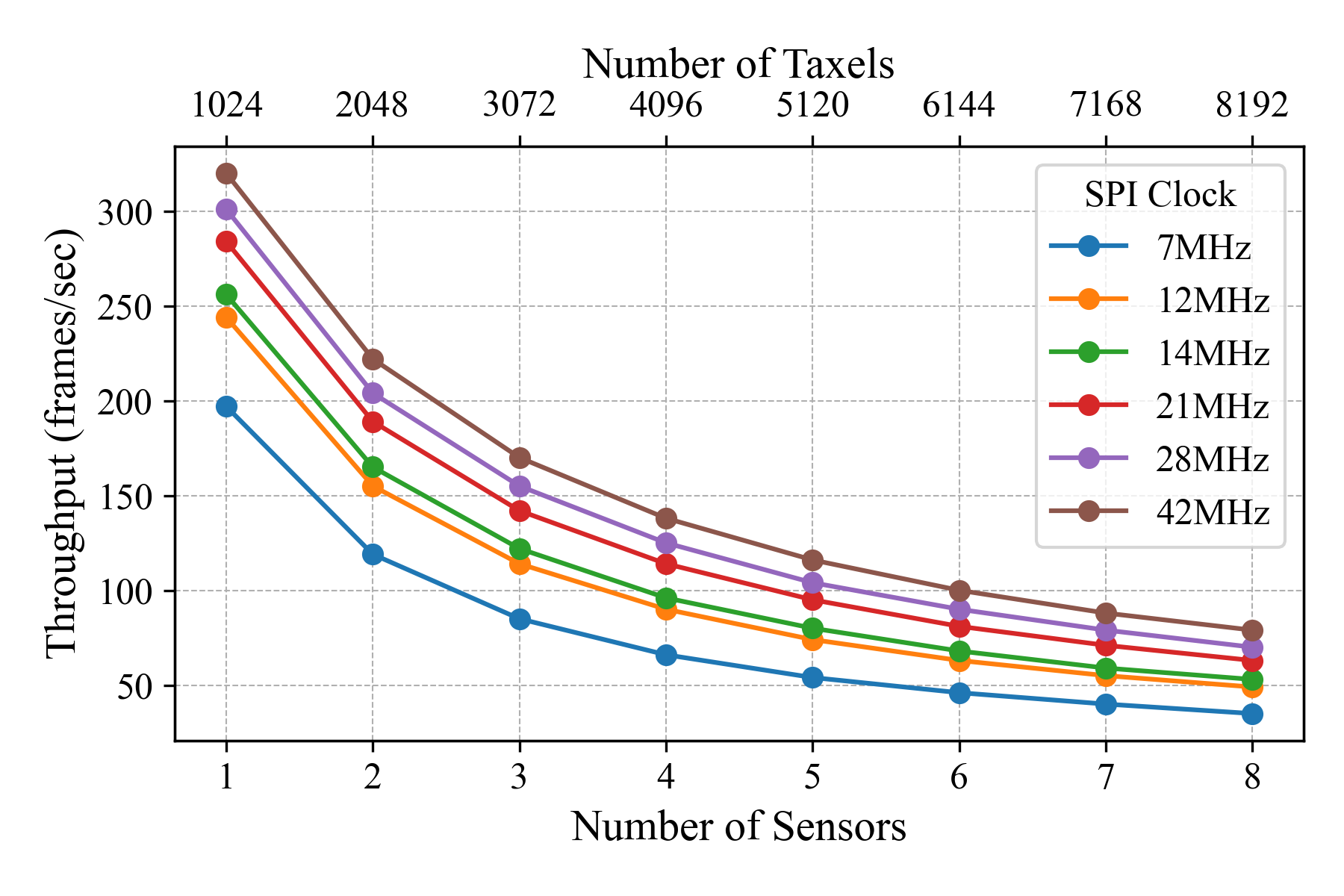}
    \caption{Frame rates as a function number of sensors on the shared SPI bus and SPI Clock frequency. As the physical length of the SPI bus increases, the SPI clock typically needs to be slower to maintain signal integrity. }
    \label{fig:frame-rates}
\end{figure}

\paragraph{Frame Rate vs Number of Sensors}
The results of this experiment are shown in Figure~\ref{fig:frame-rates}. As predicted by our timing model (Equation~\ref{eq:runtime}), the frame rate exhibits a nonlinear decay with increased sensors. Increasing the SPI clock rate decreases $t_{\text{SPI}}$. For our hardware setup the data began to degrade at clock rates past 14 Mhz, though the maximum achievable data rate is very dependent on each situation. At this frequency, the system achieved a frame rate of 256 FPS for a single sensor and maintains a frame rate of 53 FPS even when scaled to the full configuration of eight sensors. This result confirms that our architecture meets and exceeds the 50 FPS target required for real-time control at scale.

\begin{figure}
    \centering
    \includegraphics[width=0.9\columnwidth]{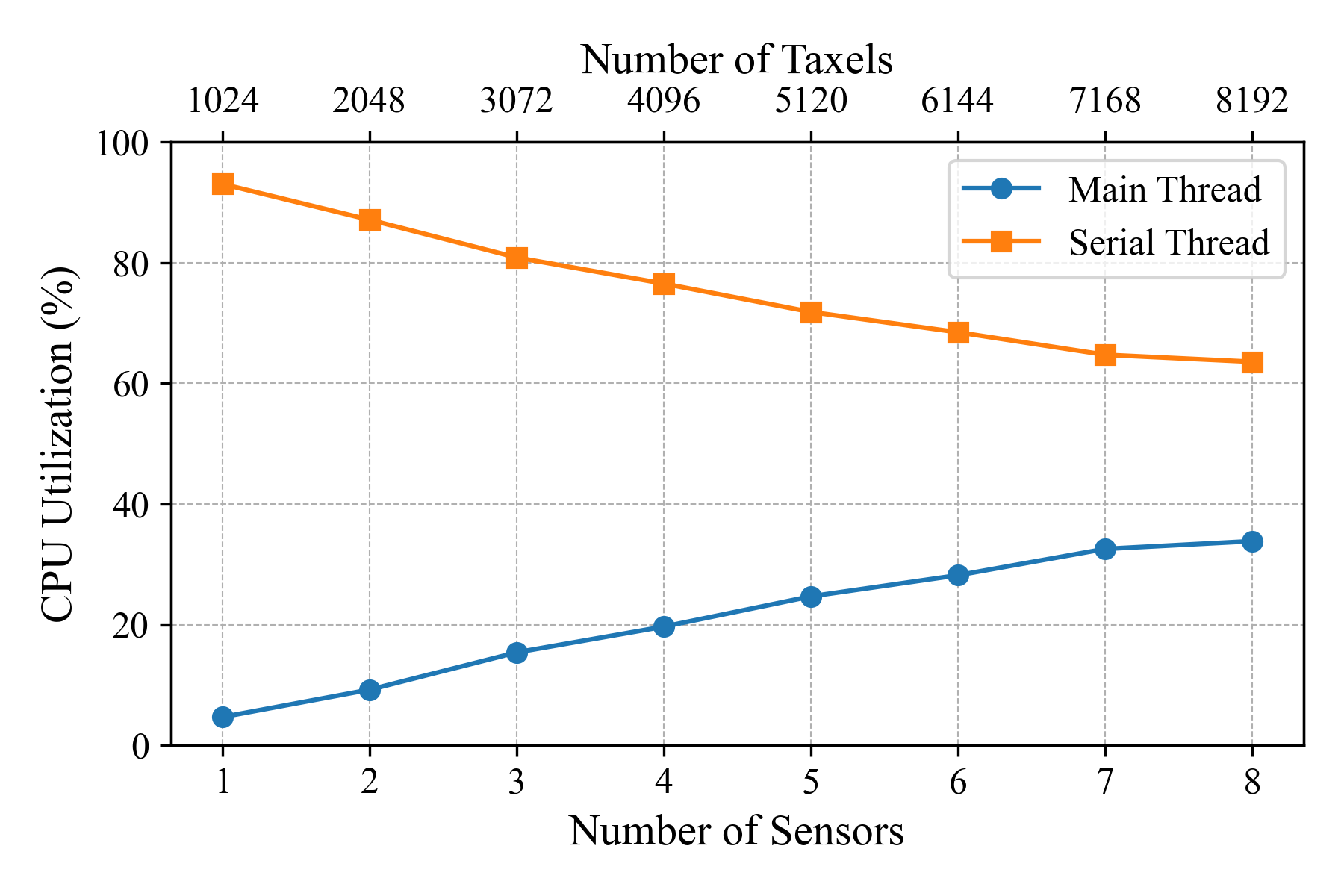}
    \caption{CPU utilization of both threads as a function of number of sensors. The main thread is communicated data over ROS and peforming filtering functions and the serial thread is responsible for reading and saving data sent from the controller.}
    \label{fig:cpu-utilization}
\end{figure}

\paragraph{CPU Load vs Number of Sensors}
The CPU utilization results, shown in Figure~\ref{fig:cpu-utilization}, reveal the performance characteristics of the host-side software. For a low number of sensors, the system is clearly I/O-bound, with the serial reader thread consuming over 90\% of a CPU core due to the high frequency of incoming data packets. As the number of sensors increases, the system frame rate decreases, which reduces the load on the reader thread. Concurrently, the processing thread's CPU load increases linearly with the volume of data it must filter per frame. While the system remains I/O-bound across the tested range, the convergence of the two lines clearly illustrates the shifting nature of the computational bottleneck from I/O handling towards data processing as the system is scaled up.

The significance of this CPU utilization analysis extends beyond simple resource measurement. By identifying the I/O handling thread as the system's bottleneck, these results provide a clear road map for future performance enhancements: efforts should be directed at optimizing the data acquisition pipeline (e.g., via a C++ implementation or lower level usb stack) rather than the already-efficient Numba-JIT-compiled data filtering algorithm. A more efficient data acquisition pipeline can either decrease CPU usage while maintaining frame rates or boost frame rates for the same CPU usage.

\paragraph{End-to-End Latency and Jitter}

\begin{figure}
    \centering
    \includegraphics[width=0.9\columnwidth]{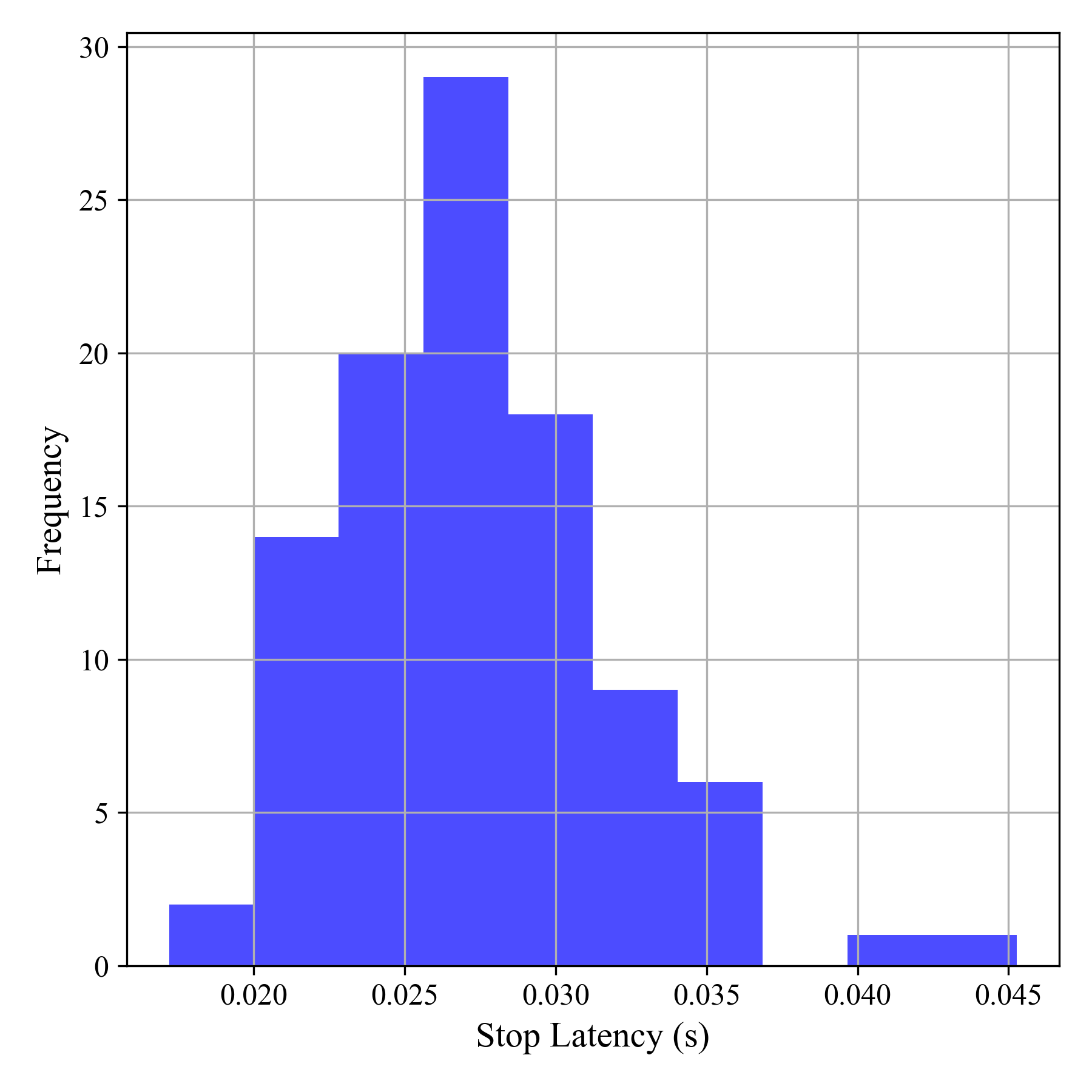}
    \caption{Peak latency histogram over 100 trials. Mean latency is 27.3 ms and jitter (standard deviation of latency) is 4.64 ms. }
    \label{fig:latency-hist}
\end{figure}

The dynamic impact tests, conducted with the $N=3$ sensor configuration, revealed a mean end-to-end latency of 27.3 ms. The latency distributions is shown in Figure~\ref{fig:latency-hist}. The jitter, calculated as the standard deviation of the latency measurements over 100 trials, was 4.64 ms. Crucially, this measured jitter is on the same order of magnitude as the calculated measurement uncertainty of ±4.5 ms arising from timestamp quantization. With a quantization error of 4.5ms, the true jitter can be approximated by the following:  $\sigma_{\text{measured}}^2 = \sigma_{\text{true}}^2 + \sigma_{\text{quantization}}^2$. Solving for $\sigma_{\text{true}}$, the estimated true jitter is about 1.13 ms. While these results are low enough for effective closed-loop control and collision detection, we attribute a portion of the latency and jitter to the host-side data handling pipeline, which is implemented in Python. Overhead from the pyserial library calls, data unpacking, and filtering steps within the Python environment all contribute to this total delay. We anticipate that both latency and jitter could be substantially reduced by reimplementing this host-side pipeline in a compiled language such as C++.

\subsection{Sensor Validation}

\begin{figure}
    \centering
    \includegraphics[width=\columnwidth]{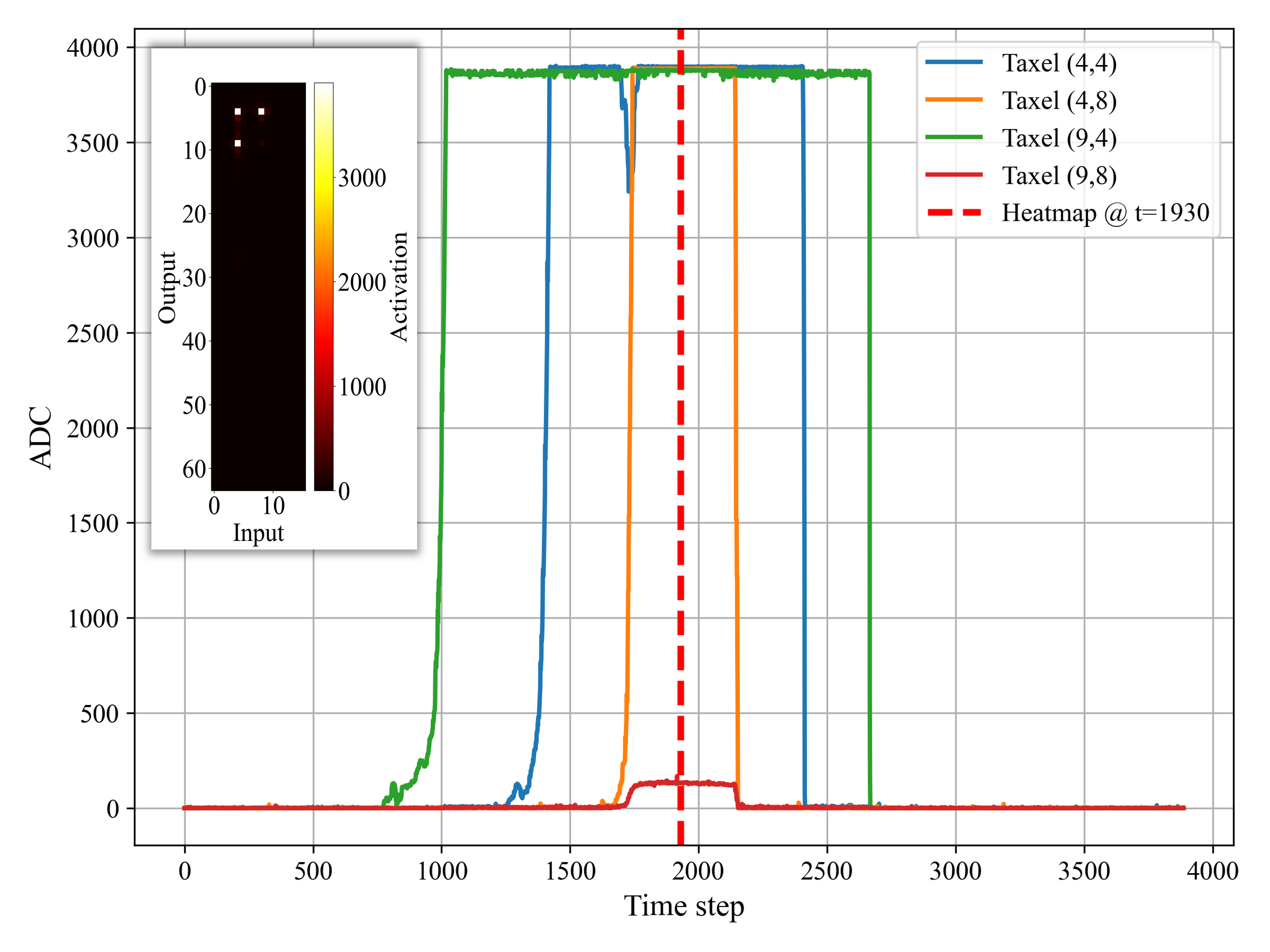}
    \caption{Crosstalk with associated heatmap. Phantom force appears on taxel (9,8) when all other taxels (shown as white pixels in top-left of figure) are pressed to create a parasitic current path. The red dotted line corresponds to the heatmap.}
    \label{fig:crosstalk-heatmap}
\end{figure}

\paragraph{Crosstalk Measurement}
The effectiveness of our crosstalk mitigation was validated in a worst-case `phantom force' test and the results are shown in Figure~\ref{fig:crosstalk-heatmap}. With three surrounding taxels ((4,4), (4,8), and (9,4)) pressed to a saturated average reading of 3886.7 (on a 12-bit ADC scale), the untouched "ghost" taxel ((9,8)) recorded an average parasitic reading of 129.8. This corresponds to a crosstalk level of only 3.3\% of the full-scale signal. This result confirms that our hardware strategy, which combines input-row guarding with output-column virtual grounding, is highly effective at preventing ghost artifacts and ensuring reliable signal fidelity during multi-touch events.

\paragraph{Adjustable Gain Demo}

\begin{figure}
    \centering
    \includegraphics[width=\columnwidth]{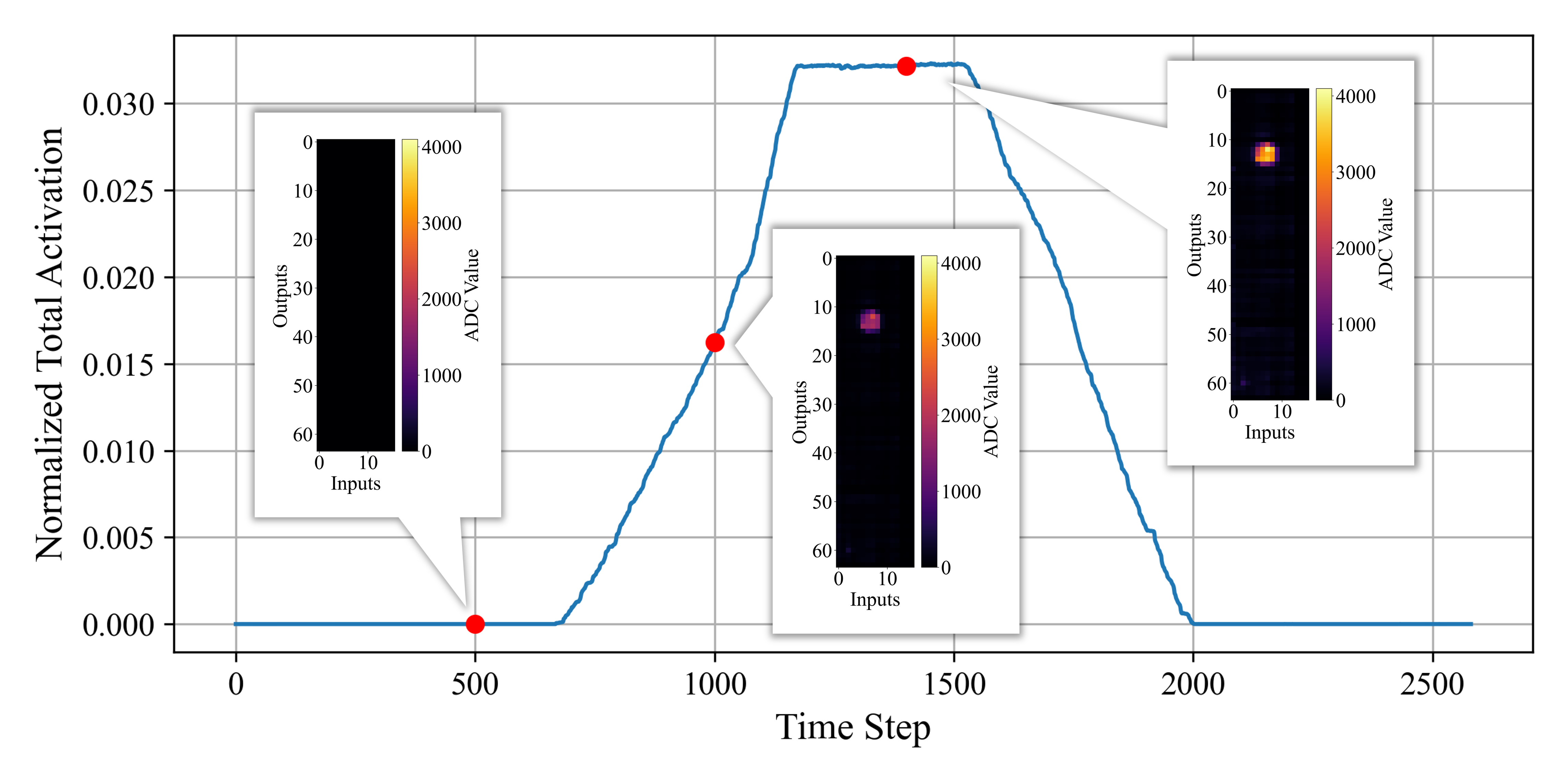}
    \caption{Adjustable sensitivity with a fixed 500g cylindrical weight sitting on a sensor across multiple taxels. Y axis is the sum of the taxel readings over the whole sensor divided by the number of taxels (1024).}
    \label{fig:adjustable-sensitivity}
\end{figure}

The functionality of the adjustable gain circuit is demonstrated in Figure~\ref{fig:adjustable-sensitivity}. With a 500g weight sitting on the sensor across multiple taxels, the output signal was initially at a low level. By adjusting the on-board potentiometer, we increased the gain of the amplifier, resulting in a significant amplification of the output signal. To report the sensor readings for all 1024 taxels as a single number, we compute the normalized total activation, which is the sum of all 1024 taxel readings divided by the number of taxels. This feature is critical for practical deployment, as it allows each sensor's sensitivity to be tuned to compensate for variations in fabric materials or to optimize the sensor's dynamic range for a specific task, such as detecting either very light touches or high forces.

\subsection{Whole-Body Tactile Sensing Ablation Study}
\begin{figure}
    \centering
    \includegraphics[width=\columnwidth]{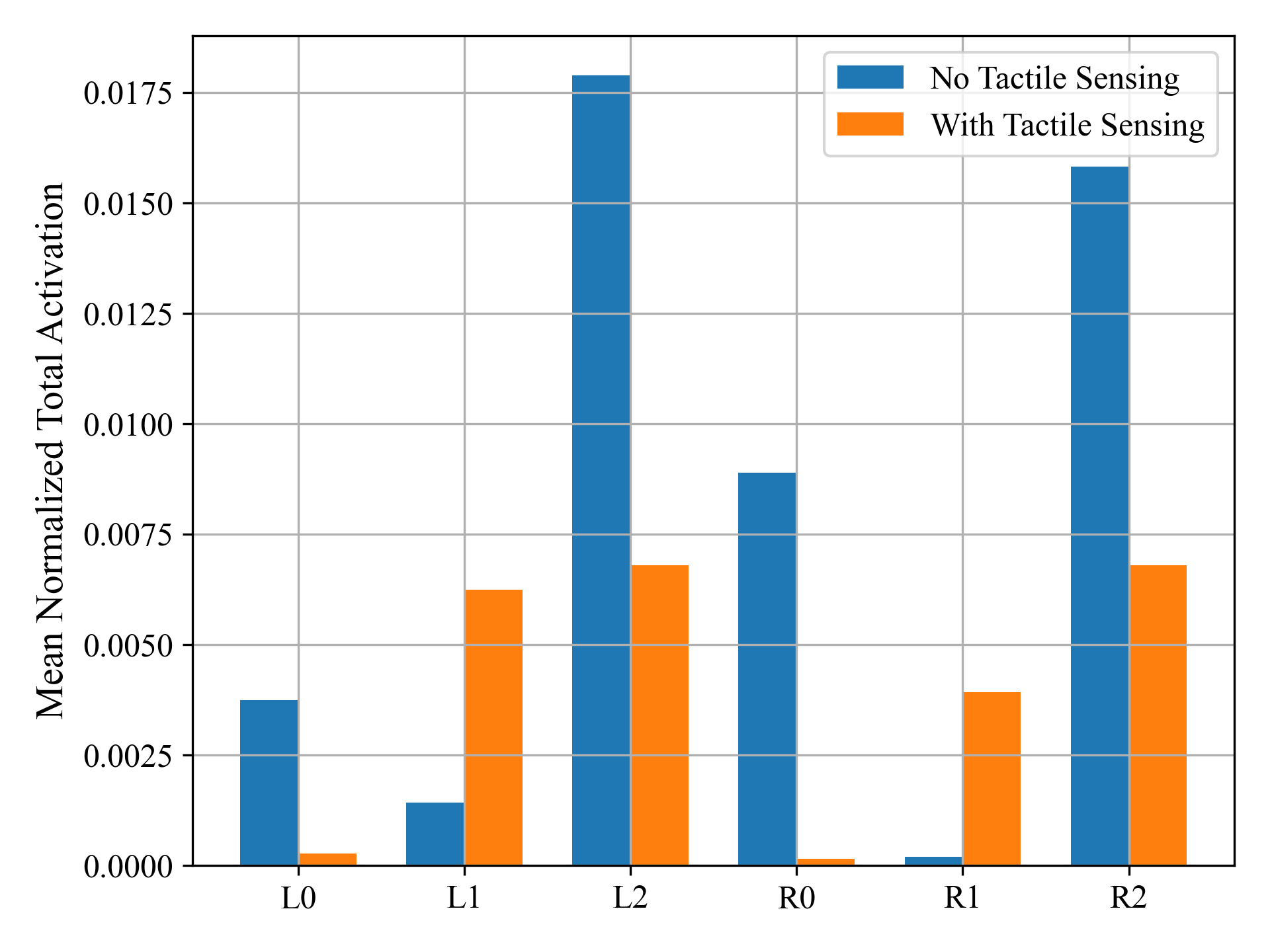}
    \caption{Normalized Total Activation for each of the six sensors during whole body grasp. Values are averages in time over the whole grasp period.}
    \label{fig:tactile-ablation}
\end{figure}

\begin{figure}
    \centering
    \includegraphics[width=\columnwidth]{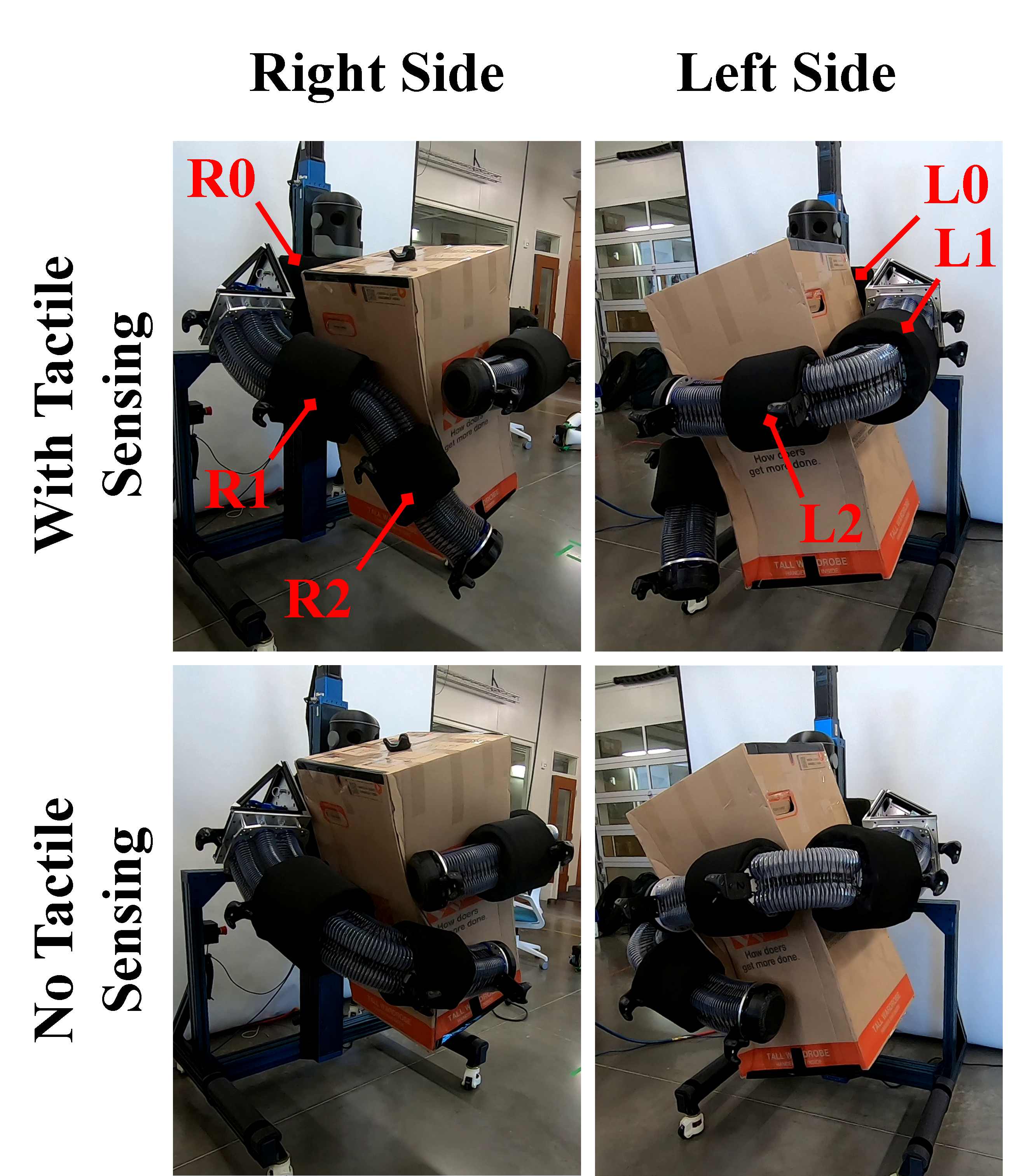}
    \caption{Whole-body grasping of large empty box. Grasping without tactile feedback increases box deformation significantly. }
    \label{fig:grasp-comparison}
\end{figure}

We validated the practical value of our system in a final ablation study, with quantitative and qualitative results shown in Figures~\ref{fig:tactile-ablation} and~\ref{fig:grasp-comparison} respectively. A video of this experiment is available online\footnote{\url{https://youtu.be/1ap56IR3Y6U}}. Without tactile feedback, the robot executes a precomputed grasping trajectory (with a set sequence of pressures commanded to each joint) with no regard for the object. The empty box clearly deforms significantly as the arms crush it during the grasp. This resulted in very high activations on sensors L2 and R2 with some contact at L0 and R0 which are located on the chest of the robot. 

In contrast, with whole-body tactile sensing enabled, the grasping trajectory was modulated in response to excessively high tactile signals (note the difference in arm configurations in Figure~\ref{fig:grasp-comparison}). The peak activations on L2 and R2 were reduced and contact patches were redistributed across more sensorized surfaces, evidences by the increased activations on L1 and R1. Activations on L0 and R0 decreased because the arms did not pull the object into the chest as strongly.

This successful demonstration provides a holistic validation of our entire system. It proves that our architecture can scale to whole-body sensing (over 6000 taxels), can be used for real-time tactile feedback control, and is robust enough to function reliably when integrated directly onto the robot's arm near sources of electrical interference like pneumatic valves and other electronics.

\section{Conclusion and Future Work}
\label{sec:conclusion}

In this paper, we have presented the design, validation, and implementation of a scalable system for whole-body robotic tactile sensing. Our work addressed the critical challenges of wiring complexity, system synchronization, and data throughput that have traditionally hindered the deployment of large-scale piezoresistive tactile sensing arrays. 

Our primary contributions include open-source sensor and electronics design\textsuperscript{\ref{repo}} that incorporates best-practices in crosstalk reduction and adjustable sensitivity. Most critically, we introduced a novel, high-throughput system architecture based on a daisy-chained SPI bus. Our experimental results demonstrated that this architecture meets the demands for real-time control, providing low latency ($<$ 30 ms), minimal jitter (1.13 ms) data from up to 8192 taxels at update rates exceeding 50 Hz. The final whole-body grasping demonstration using two buses with 3 sensors each (for a total of 6000 taxels) demonstrates real-world deployment capabilities of the architecture. 

While this work improves the scalability of tactile sensing arrays and enables whole-body sensing, there are various clear avenues for future research and improvements. Our performance characterization identified the host-side I/O handling as a primary bottleneck; re-implementing this in a complied language like C++ would significantly reduce end-to-end latency and jitter. Future work could also include a more rigorous quantitative analysis of the system's EMI immunity under controlled noise conditions to fully characterize its operational limits. Developing methods to characterize the sensors and calibrate several thousand taxels values to provide force measurements is also an important  area to investigate. Furthermore, developing more sophisticated control strategies (such as whole-body impedance control) would lead to much more intelligent whole-body manipulation capabilities. Our whole-body tactile sensing system could be used to gather large scale contact data on human or robot subjects for developing machine learning models for higher-level perception, such as large object classification or tactile pose estimation.

Ultimately, we provide a robust, accessible, and well-characterized platform that will facilitate future research into advanced whole-body robotic manipulation, human-robot interaction, and the development of more physically intelligent machines.



\section*{Acknowledgment}
This material was supported by the National Science Foundation under Grant no. 1935312.

\ifCLASSOPTIONcaptionsoff
  \newpage
\fi



\bibliographystyle{IEEEtran}
\bibliography{IEEEabrv, main}
%



%


\begin{IEEEbiographynophoto}{Curtis C. Johnson} completed his Bachelors degree in Mechanical Engineering at Brigham Young University in 2020. He is pursuing his Ph.D. in robotics under Dr. Marc Killpack in the Robotics and Dynamics Laboratory. His research interests lie at the intersection of tactile perception, model predictive control, adaptive control, and machine learning for robotics. Curtis is focused on developing control and planning algorithms for autonomous whole-body manipulation of objects with soft robots.
\end{IEEEbiographynophoto}


\begin{IEEEbiographynophoto}{Daniel Webb} completed his Bachelors Degree in Mechanical Engineering from BYU in 2025. 
He currently works as a Propulsion Engineer for SpaceX in Los Angeles, CA, USA.

\end{IEEEbiographynophoto}
\begin{IEEEbiographynophoto}{David Hill} has a Bachelors degree Electrical Engineering, with minors in both Computer Science and Mathematics at BYU. He recently completed a Masters Degree in Robotic Systems Development at Carnegie Mellon University. 
\end{IEEEbiographynophoto}
\begin{IEEEbiographynophoto}{Dr. Marc Killpack} completed his Ph.D. in Robotics from the Georgia Institute of Technology and joined BYU as an Associate Professor in December of 2013. His areas of expertise include soft robotics, human-robot interaction, controls, mechanics, and perception for robotics and other automated systems. His current research interests relate to improving modeling and control for robot manipulation in unstructured and difficult environments. This includes applications related to using soft robots for search and rescue, disaster response and human-robot interaction. While at Georgia Tech in the Healthcare Robotics Lab (HRL), Marc worked on projects including sensing and control for mobile robot bases, automating learning from robot grasping, manipulation around and interacting with human subjects, and control of a robot arm in cluttered and unmodeled environments. Prior to joining HRL, he completed Masters degrees in Mechanical Engineering from both Georgia Tech and AM Paris Tech (formerly ENSAM) in Metz, France. In 2007, Marc graduated with a Bachelor's of Science in Mechanical Engineering from Brigham Young University.
\end{IEEEbiographynophoto}




\end{document}